\documentclass[10pt]{article}
\usepackage[letterpaper]{geometry}
\usepackage{hicss}
\usepackage{times}
\usepackage[none]{hyphenat}
\usepackage{url}
\usepackage{latexsym}
\usepackage{minted}
\usepackage{indentfirst}
\usepackage{graphicx}
\usepackage{amsmath}
\graphicspath{{images/}}
\usepackage[
    style=apa,
  ]{biblatex}
\usepackage{microtype}
\usepackage{booktabs}
\usepackage{amsmath,amsfonts,amssymb}

\title{Target-Model Skill in Explainable AI: Evaluating Local Surrogate Fidelity and Perturbation Validity}

\author{Christopher Burger \\
 Pelican Quantitative \\
 {chris@pelicanqc.com}}

\date{}

\begin{document}
\maketitle
\begin{abstract}

Local surrogate explanations approximate a predictive model's behavior near a particular input, but their fidelity depends on both the evaluation region and the data used to construct the approximation. We examine three aspects of local surrogate evaluation: predictive agreement with the target model, the validity of generated perturbations, and the sensitivity of explanations to sampling procedures.

In a synthetic classification experiment, mean squared probability disagreement was 0.1094 near the explanation targets, compared with 0.0536 over the full neighborhoods. Relative to a constant reference forecast, mean Target-Model Skill was 0.258 near the targets and 0.742 overall. In experiments using numerical features from the UCI Adult dataset, independent and covariance-preserving Gaussian sampling violated specified physical constraints in 72.86\% and 73.20\% of generated observations, respectively. Across 250 Adult explanation targets, changes in perturbation sampling altered local surrogate coefficients without comparable changes in held-out predictive disagreement.

We measure target-model fidelity as agreement under a specified evaluation distribution and define Target-Model Skill as an improvement over an explicit reference forecast. For probability-valued surrogates, squared disagreement has an interpretation through expected Brier-score regret. These model-agnostic comparisons should be accompanied by separate checks of perturbation validity and explanation sensitivity.
\end{abstract}

\subsubsection*{Keywords:}

Local Surrogate Models, Probability Leakage, Target-Model Skill, Scoring Rules, XAI

\section{Introduction}
A local explanation makes a claim about how a predictive model behaves near a particular input. When the model is difficult to interpret directly, post-hoc explainable AI (XAI) methods can help us examine that behavior \parencite{app14198884,hassija2024interpreting}. The result depends in part on how we construct the explanation. Methods such as LIME \parencite{LIME_Ribeiro} and SHAP \parencite{SHAP_Lundberg} can change with perturbation and other methodological choices \parencite{wang2021suppressmanaginginputuncertainty,make6020055,L_fstr_m_2024}. The question is not simply whether an explanation looks plausible, but whether it represents the target model in the region we intend to examine.

We focus on \textit{local predictive surrogates} because this question can be tested directly. A surrogate fits a simpler, inspectable model to the predictions of a more complex target near an input of interest. We can then query the surrogate and target on the same inputs and measure their agreement. Such a comparison requires a specified evaluation region: a surrogate may agree across most of a sampled neighborhood yet perform poorly near the target input, and the perturbations used for fitting may not resemble the intended population. We examine these issues through three experiments with synthetic and tabular data. We distinguish target-model fidelity (agreement with the model being explained) from Target-Model Skill (TMS), which compares that fidelity with an explicit reference predictor. Perturbation validity and sensitivity to sampling choices remain separate diagnostics.

\section{Background and Evaluation Framework}

Prior studies have documented instability, adversarial manipulation, and problems arising from feature dependence in post-hoc explanations \parencite{burger2023explanations,ivankay2022fooling,slack202fooling,alvarezmelis2018robustness,pmlr-v238-hill24a}. Other work has examined how to represent and quantify uncertainty in explanations \parencite{10.1145/3583558,chiaburu2025uncertaintypropagationxaicomparison,reliable:neurips21,decker2025improvingperturbationbasedexplanationsunderstanding}. Our focus is narrower: we evaluate how closely a local predictive surrogate reproduces the target model, using queries rather than knowledge of the target's architecture, parameters, or training procedure. We use LIME in one experiment, but the evaluation itself does not depend on that implementation.

\subsection{Local Explanations as Approximations}

We distinguish three properties of local explanations. \textit{Fidelity} is agreement between the surrogate and the target model; \textit{stability} concerns changes in the explanation under repeated or modified procedures; and \textit{uncertainty} concerns what remains unknown about the explanation or its predictions. Our focus is fidelity, especially how it depends on the evaluation region and the perturbations used to fit the surrogate. Because the surrogate aims to reproduce the target rather than predict observed outcomes independently, it may faithfully reproduce the target's errors or spurious relationships. Fidelity and real-world predictive accuracy must therefore be assessed separately.

\subsection{Target-Model Fidelity} \label{sec:fidelity}

We use \textit{target-model fidelity} for agreement between a surrogate and its target under a specified evaluation distribution. For a binary classifier, let the target's predicted probability be

\[
f(x)=P_T(Y=1\mid X=x)
\]

Here \(Y=1\) denotes the positive class and \(f(x)\) is the target's predicted probability for that class at input \(x\). Let \(x_0\) be the input we want to explain, \(g_{x_0}(x)\) the surrogate fitted for that input, and \(\mu_{x_0}\) the distribution of inputs over which we want to assess agreement. For probability-valued surrogates, we measure disagreement by

\[
D_{\mu_{x_0}}(g,f)
=
\mathbb{E}_{X\sim\mu_{x_0}}
\left[
\left(g_{x_0}(X)-f(X)\right)^2
\right]
\]

Here \(X\) is an input drawn from \(\mu_{x_0}\), \(g_{x_0}(X)-f(X)\) is the difference between the surrogate and target probabilities, and \(\mathbb E\) averages the squared difference over that distribution. In a finite evaluation sample, we estimate this average by replacing the expectation with a sample mean. Lower disagreement indicates greater fidelity under the chosen distribution, which need not be the distribution used to fit the surrogate. We call $D_{\mu_{x_0}}(g,f)$ a \textit{fidelity loss}, not a skill score: it needs no reference predictor and is smaller when agreement is better. It requires target predictions but not access to the target's internals, and it does not measure accuracy against observed outcomes.

\subsection{Probabilistic Scoring Rules and Target-Model Skill} \label{sec:tms}

Proper scoring rules provide a useful basis for evaluating probabilistic predictions. A scoring rule is \textit{proper} when reporting the true probability distribution minimizes expected loss. It is \textit{strictly proper} when that minimum is unique \parencite{Gneiting01032007}.

For a binary outcome \(y\in\{0,1\}\), the Brier loss is the squared difference between the forecast probability \(p\) and the observed outcome:

\[
S(p,y)=(p-y)^2
\]

To apply the score to surrogate fidelity, treat the target probability \(f(x)\) as the reference Bernoulli distribution at \(x\). The difference in expected Brier loss between the probability-valued surrogate and the target is

\[
\begin{aligned}
&\mathbb{E}\left[S(g_{x_0}(x),Y)\mid X=x\right]-\mathbb{E}\left[S(f(x),Y)\mid X=x\right]\\
&=\left(g_{x_0}(x)-f(x)\right)^2
\end{aligned}
\]

Squared probability disagreement is therefore the surrogate's expected Brier-score regret relative to the target model. Averaging over the evaluation distribution gives the fidelity loss in Section~\ref{sec:fidelity}. This interpretation uses the target's output as the reference forecast, not as verified ground truth. The surrogate must also report probabilities for a defined event: merely returning a number between zero and one does not, without that interpretation, establish a probabilistic forecast. For a specified binary event, a value \(p\) in that interval and its complement \(1-p\) do form a valid Bernoulli distribution, although they need not be calibrated to observed outcomes.

The terms \textit{hard labels} and \textit{soft targets} describe what the surrogate learns from during fitting. A hard label keeps only the target's class decision, \(h_T(x)=\mathbf 1[f(x)\geq 0.5]\); a soft target retains its predicted probability, \(f(x)\). For example, target probabilities of 0.51 and 0.99 both become the hard label 1, but remain different soft targets. In either case, a logistic surrogate itself outputs a probability. We therefore distinguish the target values used to fit the surrogate from the reference values used to score it. We examine both training choices in Experiment~1.

In forecasting, skill scores express performance relative to a reference forecast \parencite{wilks2011statistical}. We reserve the term \textit{Target-Model Skill} (TMS) for this reference-relative comparison. Given an explicitly chosen predictor \(r\), define

\[
\operatorname{TMS}_{\mu}(g;r)
=
1-
\frac{D_{\mu}(g,f)}
{D_{\mu}(r,f)}
\]

when $D_{\mu}(r,f)>0$. Here $r$ is a reference predictor evaluated against the same target and on the same input distribution as $g$. A TMS of one means exact agreement with the target, zero means the same disagreement as the reference, and a negative value means worse agreement than the reference. Unlike the fidelity loss $D_{\mu}$, TMS is higher when reference-relative performance is better.

We report raw fidelity loss in all predictive-surrogate experiments and TMS where an explicit reference predictor is available (Experiment~1). These are distinct measures; neither establishes calibration against observed outcomes. Calibration concerns whether events assigned a probability \(p\) occur with frequency \(p\) among comparable forecasts, whereas our scores compare one model's outputs with another's. Nor do these scores quantify \textit{epistemic uncertainty}, which concerns uncertainty due to incomplete knowledge or limited data: a single disagreement value does not indicate how much the fitted explanation might change with new data or repeated sampling. Finally, a score computed on a set of inputs cannot establish that those inputs represent the intended population. We assess their validity separately.

\subsection{The Role of the Evaluation Neighborhood} \label{sec:neighborhood}

Local surrogate methods often construct explanations by querying the target at perturbed inputs. Let \(Q_{x_0}\) denote the distribution from which the explainer draws those inputs; it is not necessarily an estimate of the observed data distribution. The surrogate is fitted using samples from \(Q_{x_0}\), but we may wish to evaluate it under a different distribution \(\mu_{x_0}\). We call these the \textit{fitting distribution} and \textit{evaluation distribution}, respectively. This distinction matters both when perturbations differ from the inputs the explanation is intended to represent and when an aggregate score masks variation across regions.

Suppose we partition an evaluation neighborhood into disjoint regions \(A_1,\ldots,A_K\). The overall squared disagreement can be written as

\[
D_\mu(g,f)
=
\sum_{k=1}^{K}
\mu(A_k)\,
D_{\mu(\cdot\mid A_k)}(g,f)
\]

The overall error is a weighted average of regional errors, so regions receiving more probability under the evaluation distribution contribute more to the score. A surrogate may consequently achieve low aggregate error while performing poorly in a small region surrounding the explanation target.

We therefore evaluate agreement across the neighborhood, within specified regions, and at the target input itself. These evaluations answer different questions. No single neighborhood is appropriate for every explanatory objective. An explanation intended to describe behavior among plausible observations may require a different evaluation distribution from one intended to probe hypothetical changes in input features.

\subsection{Perturbation Validity and Probability Leakage}

The perturbation distribution determines which target-model behaviors the surrogate observes during fitting. Two sampling problems matter here. In \textit{distributional mismatch}, a generator produces observations that are unusual under the intended data distribution, for example by sampling dependent features independently. Such observations may be unlikely without being impossible. In a \textit{support violation}, the generator assigns probability to inputs excluded by known constraints, such as negative values for a necessarily nonnegative feature. The latter connects to \textit{probability leakage}: probability assigned to outcomes ruled out by the available evidence \parencite{briggs2012probabilityleakage,briggs2016uncertainty}.

Let $\mathcal C\subseteq\mathcal X$ denote the set of admissible inputs under explicitly stated constraints. We define the probability leakage of a perturbation generator \(Q\) relative to this set as

\[
L(Q;\mathcal C)
=
Q(X\notin\mathcal C)
\]

For $N$ generated observations, we estimate this quantity using

\[
\widehat L
=
\frac{1}{N}
\sum_{i=1}^{N}
\mathbf 1[x_i\notin\mathcal C]
\]

This definition concerns the perturbation generator, not necessarily the target or surrogate model. It also depends on the constraints we specify. We distinguish physical restrictions, dataset-selection criteria, and recording conventions: an observation excluded from a study population may still be physically possible. We therefore measure explicit violations separately from statistical distributional mismatch.

A generator can preserve correlations while violating constraints or respect every stated constraint while producing unusual feature combinations. Neither property alone determines surrogate fidelity. We therefore report fidelity, perturbation validity, and sensitivity to the explanation procedure as separate diagnostics rather than combining them into one reliability score.

\section{Methodology}

We conducted three experiments. The first examines fidelity at different distances from an explanation target; the second tests whether perturbations preserve feature relationships and input constraints; and the third measures how sampling choices affect surrogate coefficients and fidelity. Throughout, we distinguish the fitting distribution \(Q_{x_0}\) from the evaluation distribution \(\mu_{x_0}\), as defined in Section~\ref{sec:neighborhood}. Unless stated otherwise, we assess fidelity using squared disagreement with the target model and assess perturbation validity separately. The first experiment fits logistic surrogates directly, the second studies perturbation generators, and the third uses LIME's local fitting procedure. None requires access to the target model's internals.

\subsection{Spatial Variation in Surrogate Fidelity} \label{sec:spatial_method}

To test whether aggregate scores conceal errors near the explanation target, we used a two-dimensional classification problem that reduces the complications of higher-dimensional features. We generated 2,000 observations using scikit-learn's \texttt{make\_moons} function with a noise parameter of 0.2. We trained a multilayer perceptron classifier with two hidden layers of 50 neurons each to serve as the target model. We then evaluated the target model on a $400 \times 400$ grid covering the data range, extended by 0.5 units in each direction. From this grid, we identified locations where the target model assigned a positive-class probability between 0.49 and 0.51. We randomly selected 50 of these locations without replacement as explanation targets.

For each target \(x_0\), we generated two independent sets of Gaussian perturbations:

\[
z_i=x_0+\epsilon_i,\qquad
\epsilon_i\sim\mathcal N(0,I).
\]

We used 3,000 perturbations to fit each logistic surrogate and an independent set of 4,500 to evaluate it. In the \textit{hard-label fit}, we first converted each target probability \(f(z_i)\) to the binary class decision \(h_T(z_i)=\mathbf 1[f(z_i)\geq0.5]\), then fitted ordinary logistic regression to those decisions. In the \textit{soft-target fit}, we retained \(f(z_i)\) as the training target. We fitted logistic coefficients by minimizing the expected Bernoulli log loss, with the target probability weighting the losses for classes 1 and 0, and a small quadratic coefficient penalty (\(10^{-4}\) on the squared slopes). Thus, both fitted surrogates return probabilities; the distinction is whether training uses the target's class decisions or its probability estimates. The fits use different regularization implementations, so this comparison is a robustness check rather than an isolated test of the effect of target encoding.

For each evaluation observation, we calculated its Euclidean distance \(d\) from the explanation target and divided the neighborhood into three regions:
\begin{itemize}
    \item  Inner: \(d<0.3\)
    \item Middle: \(0.3\leq d<0.7\)
    \item Outer: \(d\geq0.7\)
\end{itemize}

These radii partition the Gaussian neighborhood into an immediate target region, a middle region, and an outer region. The boundaries were specified for evaluation, not optimized, and need not suit other models or perturbation scales.

We calculated squared probability disagreement within each region and across the entire evaluation neighborhood:

\[
\widehat D_A(g,f)
=
\frac{1}{|A|}
\sum_{i\in A}
\left(g_{x_0}(z_i)-f(z_i)\right)^2
\]

where \(A\) denotes the evaluation observations in a particular region. For the binary-label fit, we also calculated the Brier score against the target's class decisions. To report reference-relative TMS, we defined a constant probability predictor for each target, $r_{x_0}(x)\equiv n_{\mathrm{fit}}^{-1}\sum_{i=1}^{n_{\mathrm{fit}}}f(z_i^{\mathrm{fit}})$, where $z_i^{\mathrm{fit}}$ are that target's fitting perturbations. We compared the hard-label-trained surrogate with this reference on identical held-out observations, computed TMS separately for each target and spatial region, and averaged the 50 scores. This TMS comparison uses target \emph{probabilities}, not hard class labels.

At each explanation target, we also measured absolute probability disagreement to distinguish pointwise from neighborhood performance. We summarized scores across 50 targets, used target-level bootstrap intervals to describe variability, and recorded the fraction of evaluation observations in each region to show how sampling weights the overall score.

\subsection{Perturbation Validity and Probability Leakage}

The second experiment tests whether generated perturbations preserve feature relationships and satisfy known input constraints. We evaluate these properties separately: a low-density observation need not be impossible, and a constraint violation can occur regardless of estimated density.

\subsubsection{Feature Dependence}

We first generated a two-dimensional bivariate normal reference distribution with unit marginal variances and correlation \(\rho=0.95\). For each explanation target, we compared two perturbation generators. The first sampled independent Gaussian noise using the reference marginal variances. The second sampled from a Gaussian distribution preserving the reference covariance matrix. Both generators used the same target coordinates and marginal scales. We examined perturbation scales of 0.25, 0.50, and 1.00, using 30 target locations and 400 perturbations per target at each scale.

Because the reference distribution was known, we could evaluate generated observations using its exact probability density. We defined a low-density threshold at the 10th percentile of log-density values from an independent calibration sample. We then calculated the proportion of generated observations falling below this threshold.

To test whether the same behavior occurred in observed data, we repeated the comparison using pairs of numerical features from scikit-learn's breast cancer dataset. We examined one strongly correlated pair (mean radius and mean perimeter) and one more weakly correlated pair (mean radius and mean texture).

We divided the observations into training, calibration, and evaluation sets containing approximately 60\%, 20\%, and 20\% of the data, respectively. We standardized the features using training-set statistics and fitted a 10-nearest-neighbor distance estimator to the training observations. We defined the anomaly threshold as the 90th percentile of nearest-neighbor distances among the calibration observations. Both perturbation generators were evaluated using this same threshold.

For each feature pair, we selected 30 evaluation targets and generated 350 perturbations per target at scales of 0.15, 0.35, and 0.70. We used paired noise draws so that the intended difference between generators was their covariance structure rather than differences in the underlying random samples.

These experiments measure whether preserving covariance changes the proportion of generated observations classified as low-density. They do not establish whether those observations are logically or physically impossible.

\subsubsection{Constraint Validity and Probability Leakage}

We examined explicit input constraints using five numerical features from the UCI Adult Census Income dataset \parencite{census_income_20}: age, education-num, capital-gain, capital-loss, and hours-per-week. All five features are recorded numerically, although several are integer-valued or concentrated at zero.

We divided the data into training and testing sets using a 75/25 split. Across ten random seeds (1--5 and 42--46), we selected 25 explanation targets per split, giving 250 targets in total. Each selection included test observation 42 and 24 additional randomly selected observations.

For each target, we generated 5,000 observations using three procedures:
\begin{enumerate}
    \item Independent Gaussian sampling, which adds independent noise scaled by each feature's training-set standard deviation.
    \item Covariance-preserving Gaussian sampling, which draws multivariate Gaussian noise using the training-set covariance matrix.
    \item Local empirical resampling, which draws observations with replacement from the 150 nearest training records in standardized feature space.
\end{enumerate}

We evaluated the generated observations against explicit constraints. Physical constraints included nonnegative age, capital gain, and capital loss, and a range of 0 to 168 hours per week. We separately checked the Adult dataset's selection criteria, including age greater than 16 and positive hours worked.

We also recorded whether generated values differed from the dataset's integer-valued representation or fell outside the observed training ranges. We did not treat these representation and range checks as equivalent to physical impossibility. Let \(\mathcal C\) denote the set of observations satisfying the physical constraints. For a generated sample \(z_1,\ldots,z_N\), we estimated probability leakage using

\[
\widehat L(Q;\mathcal C)
=
\frac{1}{N}
\sum_{i=1}^{N}
\mathbf 1[z_i\notin\mathcal C]
\]

We first checked these constraints against observed Adult records. For the independent Gaussian generator, we calculated the expected violation probability for capital gain and capital loss as a check on the simulated rates. We applied the same categories of constraints to the six-feature sampling configurations in Experiment 3. These comparisons test whether preserving covariance also respects the admissible input domain.

\subsection{Perturbation Sampling and Attribution Sensitivity}

The final experiment tests how perturbation sampling changes the explanations fitted to a fixed target model. We used six Adult features: age, hours-per-week, workclass, marital-status, occupation, and relationship. We treated the first two as numerical and encoded the remaining four as categorical.

For ten random seeds (1--5 and 42--46), we split the observations into training (60\%), calibration (20\%), and evaluation (20\%) sets. We fitted numerical standardization and categorical one-hot encoding on the training data. The target was a Random Forest with 100 trees and a minimum of three observations per leaf. We selected 25 explanation targets per split, giving 250 targets in total.

\subsubsection{Controlled Perturbation Procedures}

For each explanation target, we identified the 150 nearest training observations in the transformed feature space. We used these observations to construct a local empirical neighborhood of 1,500 samples, including the explanation target itself. We compared six sampling configurations:
\begin{itemize}
    \item \textbf{Joint reference}: Resample complete observations from the local empirical neighborhood, preserving the observed feature combinations.
    \item \textbf{Joint redraw}: Generate a second neighborhood from the same local training pool to measure ordinary sampling variation.
    \item \textbf{Numerical shuffle}: Independently permute the numerical feature columns of the reference neighborhood.
    \item \textbf{Categorical shuffle}: Independently permute the categorical feature columns.
    \item \textbf{All-feature shuffle}: Independently permute all feature columns.
    \item \textbf{Native LIME}: Use LIME's own perturbation generator as a separate comparison.
\end{itemize}

Shuffling preserves each feature's empirical marginal values while changing cross-feature relationships. For every neighborhood, we used LIME's tabular procedure to fit a locally weighted Ridge surrogate (\(\alpha=0.2\)) with all six features, 1,500 samples, and continuous-feature discretization disabled. For the controlled conditions, we replaced perturbation generation while retaining LIME's local weighting, interpretable representation, and surrogate fitting. The native configuration instead uses training-derived marginal statistics for Gaussian perturbations of numerical features and empirical frequencies for categorical features. It can therefore differ from the local empirical neighborhoods in sampling scale, population, support, and feature dependence, not merely in whether columns are shuffled. We treat native LIME as a practical configuration comparison, not an isolated test of a single sampling mechanism or an evaluation of LIME in general.

\subsubsection{Evaluation Measures}

For each explanation target, we selected the 50 nearest other observations in the held-out evaluation set and used them to assess all six surrogates against the same target model. We calculated mean squared surrogate-output disagreement on these neighbors and absolute disagreement at the explanation target.

To compare feature attributions, we represented each explanation by its six local regression coefficients. We measured the Euclidean distance between each coefficient vector and that of the joint-reference explanation:

\[
\Delta_\beta
=
\left\|
\boldsymbol\beta_{\mathrm{method}}
-
\boldsymbol\beta_{\mathrm{joint}}
\right\|_2
\]

We also recorded whether the largest absolute coefficient belonged to a different feature than in the joint-reference explanation. These comparisons measure sensitivity, not attribution correctness. Finally, we assessed perturbation validity using a 15-nearest-neighbor distance estimator. We fitted this estimator on half of the calibration observations, disjoint from both the model-training data and the empirical perturbation pool. We set the anomaly threshold at the 90th percentile of distances for the other half of the calibration observations. Neither subset overlapped with the independent evaluation observations.

We applied one detector and threshold to all configurations, including independent real observations used to establish a reference rate. We summarized disagreement, coefficient changes, feature-rank changes, and anomaly rates across the 250 targets. Pairing the methods at each target allowed us to distinguish changes in the explanation from changes in predictive agreement.

\section{Experimental Results}

We report spatial fidelity, perturbation validity, and attribution sensitivity separately. For probability-valued surrogates we use squared probability disagreement and, in Experiment~1, reference-normalized TMS; for the unbounded Ridge surrogates we use squared output disagreement. Neither error measure nor the normalized skill score substitutes for constraint checks, anomaly rates, or changes in attribution coefficients.

\subsection{Spatial Variation in Surrogate Fidelity}

\begin{figure*}[t]
\centering
\includegraphics[width=\textwidth]{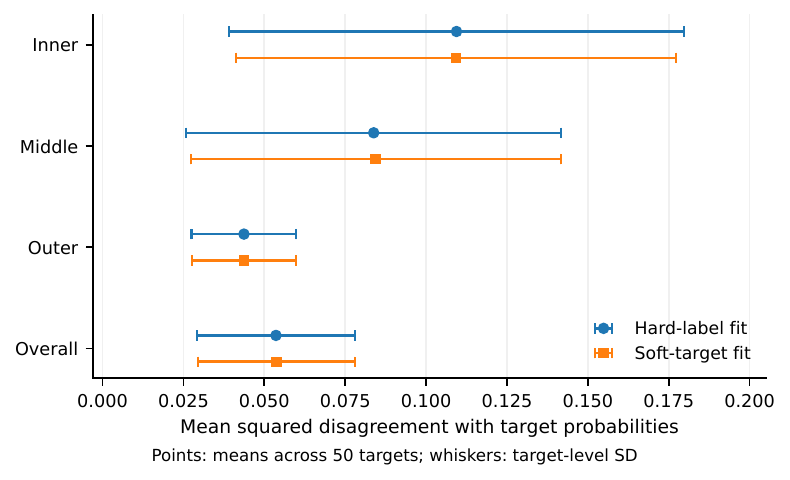}
\caption{Held-out squared probability disagreement by region for hard-label and soft-target logistic surrogates. Points show means across 50 explanation targets; whiskers are 95\% bootstrap intervals obtained by resampling targets. Lower values indicate better agreement.}
\label{fig:spatial_fidelity}
\end{figure*}

The first experiment found substantial spatial variation in fidelity on independent evaluation observations (Figure~\ref{fig:spatial_fidelity}). Table~\ref{tab:spatial_fidelity} summarizes mean scores across 50 targets near the MLP decision boundary.

\begin{table}[t]
\centering
\caption{Held-out error across spatial regions, averaged over 50 explanation targets. Brier is scored against the target's binary class decisions for the hard-label fit; the hard- and soft-fit columns give squared disagreement with target probabilities. Lower values indicate closer agreement.}
\label{tab:spatial_fidelity}
\small
\setlength{\tabcolsep}{5.5pt}
\begin{tabular}{lccc}
\toprule
Region & Brier & Hard MSE & Soft MSE \\
\midrule
Inner ($d<0.3$)       & 0.2025 & 0.1094 & 0.1093 \\
Middle ($0.3\leq d<0.7$) & 0.1168 & 0.0838 & 0.0844 \\
Outer ($d\geq0.7$)    & 0.0589 & 0.0437 & 0.0437 \\
Overall               & 0.0753 & 0.0536 & 0.0537 \\
\bottomrule
\end{tabular}
\end{table}

The hard-label fit had a mean Brier score of 0.2025 in the inner region versus 0.0589 in the outer region when scored against the target's binary decisions. Evaluating this same surrogate against target probabilities gave squared disagreements of 0.1094 and 0.0437, respectively. The soft-target fit, which retains the target's probability information during training, had nearly the same inner-region disagreement (0.1093). At the exact explanation targets, mean absolute probability disagreement was 0.1923 for the hard-label fit and 0.1875 for the soft-target fit.

The Gaussian evaluation distribution placed only 4.3\% of samples in the inner region, 17.4\% in the middle, and 78.3\% in the outer region. These weights yielded an overall hard-label-fit probability disagreement of 0.0536, approximately half the inner-region value. The paired inner-minus-overall difference averaged 0.0558, with a descriptive 95\% target-bootstrap interval of [0.04212, 0.06948].

Relative to the constant target-probability reference specified in Section~\ref{sec:spatial_method}, the hard-label-trained surrogate had mean TMS values of 0.2583 (inner), 0.5803 (middle), 0.7949 (outer), and 0.7419 (overall). We calculated each target's TMS before averaging, rather than normalizing the pooled errors in Table~\ref{tab:spatial_fidelity}. Thus, both the raw losses and reference-relative skill indicate weaker agreement near the target; the raw-error pattern also persisted under soft-target fitting.

\subsection{Perturbation Validity and Probability Leakage}

\subsubsection{Feature Dependence and Low-Density Sampling}

We first evaluated feature dependence. The synthetic experiment used a bivariate normal distribution with correlation \(\rho=0.95\). Independent Gaussian sampling and covariance-preserving Gaussian sampling produced substantially different proportions of low-density observations (Table~\ref{tab:synthetic_sampling}).

\begin{table}[t]
\centering
\caption{Low-density perturbations under a bivariate Gaussian reference distribution with correlation $\rho=0.95$.}
\label{tab:synthetic_sampling}
\small
\setlength{\tabcolsep}{4.5pt}
\begin{tabular}{rrr}
\toprule
Scale & Independent (\%) & Covariance (\%) \\
\midrule
0.25 & 10.63 & 0.00 \\
0.50 & 38.34 & 1.21 \\
1.00 & 66.78 & 17.70 \\
\bottomrule
\end{tabular}
\end{table}

\begin{figure*}[t]
\centering
\includegraphics[width=0.96\textwidth, height=190pt, keepaspectratio]{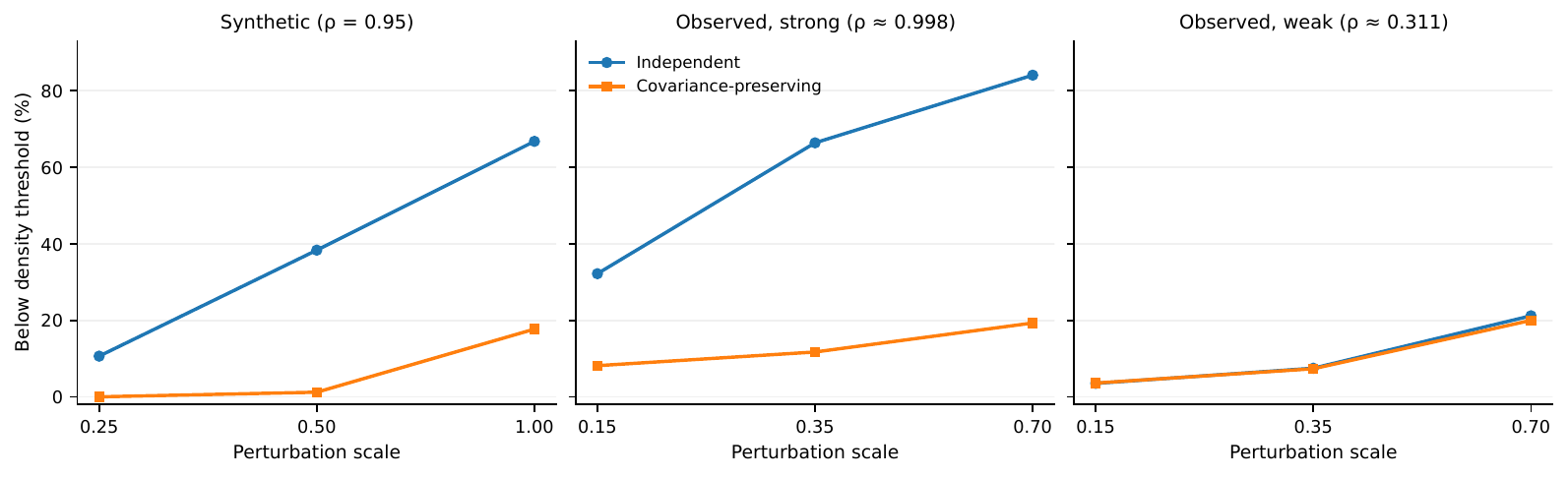}
\caption{Low-density rates under independent and covariance-preserving perturbations. Panels show a known bivariate Gaussian reference distribution, a strongly correlated observed feature pair, and a weakly correlated observed pair. The reference criteria differ between the synthetic and observed-data tests.}
\label{fig:feature_dependence}
\end{figure*}

Held-out reference observations had a 9.76\% low-density rate. At perturbation scale 0.50, independent and covariance-preserving sampling yielded rates of 38.34\% and 1.21\%, respectively. At scale 1.00, the rates were 66.78\% and 17.70\%. We also examined strongly and weakly correlated real-data features (Figure~\ref{fig:feature_dependence}).

\begin{table}[t]
\centering
\caption{Low-density perturbations for numerical feature pairs from the breast-cancer dataset at scale 0.35.}
\label{tab:real_sampling}
\small
\setlength{\tabcolsep}{4.5pt}
\begin{tabular}{lrr}
\toprule
Feature pair & Independent (\%) & Covariance (\%) \\
\midrule
Strong correlation & 66.40 & 11.72 \\
Weak correlation & 7.48 & 7.34 \\
\bottomrule
\end{tabular}
\end{table}

At scale 0.35, independent sampling yielded a 66.4\% low-density rate for mean radius and mean perimeter (\(\rho\approx0.998\)), compared with 11.7\% under covariance-preserving sampling (Table~\ref{tab:real_sampling}). For mean radius and mean texture (\(\rho\approx0.31\)), the rates were 7.48\% and 7.34\%. The contrast was pronounced for strongly dependent features but negligible for the weaker pair.

The low-density measurements do not establish that the generated observations are impossible. They indicate disagreement between the perturbation distribution and the reference distribution under the chosen density criterion.

\subsubsection{Probability Leakage}

In the five-feature Adult experiment, independent and covariance-preserving Gaussian sampling violated tested physical constraints in 72.865\% and 73.204\% of generated observations, respectively. Most violations involved negative values for capital gain or capital loss, which must be nonnegative. The local empirical bootstrap produced no violations of these constraints.

These rates follow from the Gaussian generator's support. If both capital gain and capital loss are zero at the target, independently sampled Gaussian noise with positive variance makes each negative with probability 0.5, so at least one becomes negative with probability 0.75. Preserving covariance did not solve this problem: a multivariate Gaussian still assigns probability outside the admissible domain. The six-feature mixed-data comparison yielded a different pattern.

\begin{table}[t]
\centering
\caption{Constraint and representation checks for native LIME perturbations using six mixed Adult features. Categories can overlap.}
\label{tab:mixed_constraint_checks}
\small
\setlength{\tabcolsep}{4pt}
\begin{tabular}{lr}
\toprule
Check & Flagged (\%) \\
\midrule
Physical constraint violation & 0.211 \\
Cohort-selection violation & 4.542 \\
Non-integer numerical value & 99.933 \\
Outside observed training range & 5.332 \\
\bottomrule
\end{tabular}
\end{table}

Under the native LIME sampling configuration (Table~\ref{tab:mixed_constraint_checks}), 0.211\% of generated observations violated the tested physical constraints, while 4.542\% fell outside the Adult dataset's age and working-hours selection criteria. The expected physical-violation rate under the Gaussian sampler was 0.207\%, close to the observed 0.211\%. Other discrepancies were more frequent: 99.93\% of perturbations contained non-integer numerical values and 5.33\% fell outside observed training ranges. These are representation or range differences, not necessarily physical impossibilities; cohort-selection violations likewise need not be physically invalid. None of the five empirically resampled or shuffled configurations violated the tested physical constraints.

\begin{figure*}[t]
\centering
\includegraphics[width=0.97\textwidth]{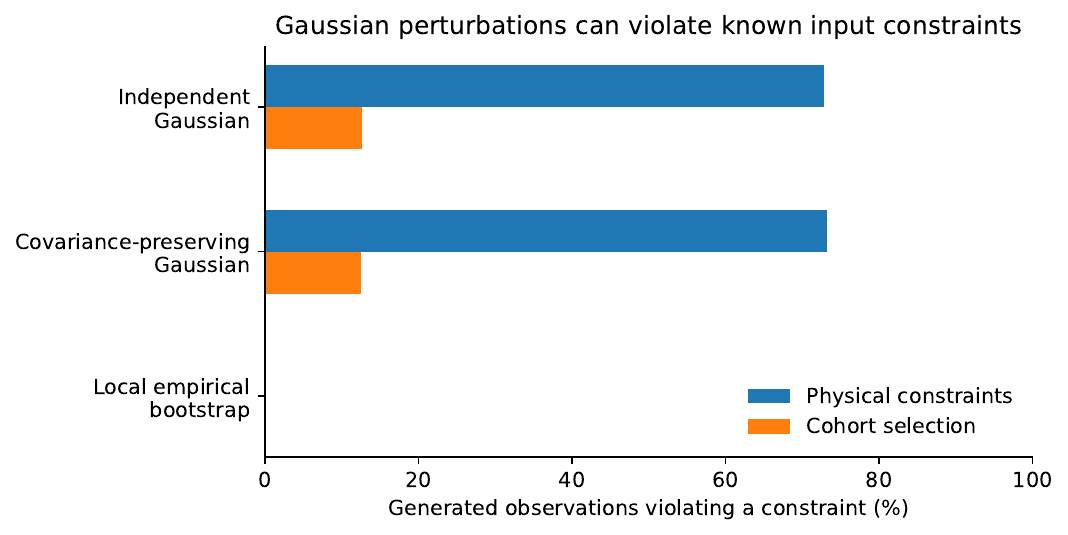}
\caption{Constraint-violation rates for the five-numerical-feature Gaussian and empirical generators (left) and the six-feature mixed-data generator comparison (right). The panels use different vertical scales and feature sets; cohort-selection violations are not necessarily physically impossible.}
\label{fig:probability_leakage}
\end{figure*}

Figure~\ref{fig:probability_leakage} shows the contrast in constraint violations. The five-feature Gaussian generators produced many inadmissible inputs, while mixed-feature native LIME produced few hard violations despite a higher anomaly rate under a separate distributional check. Preserving covariance, satisfying feature constraints, and reproducing a joint distribution are distinct properties.

\subsection{Perturbation Sampling and Attribution Sensitivity}

\begin{figure*}[t]
\centering
\includegraphics[width=0.98\textwidth]{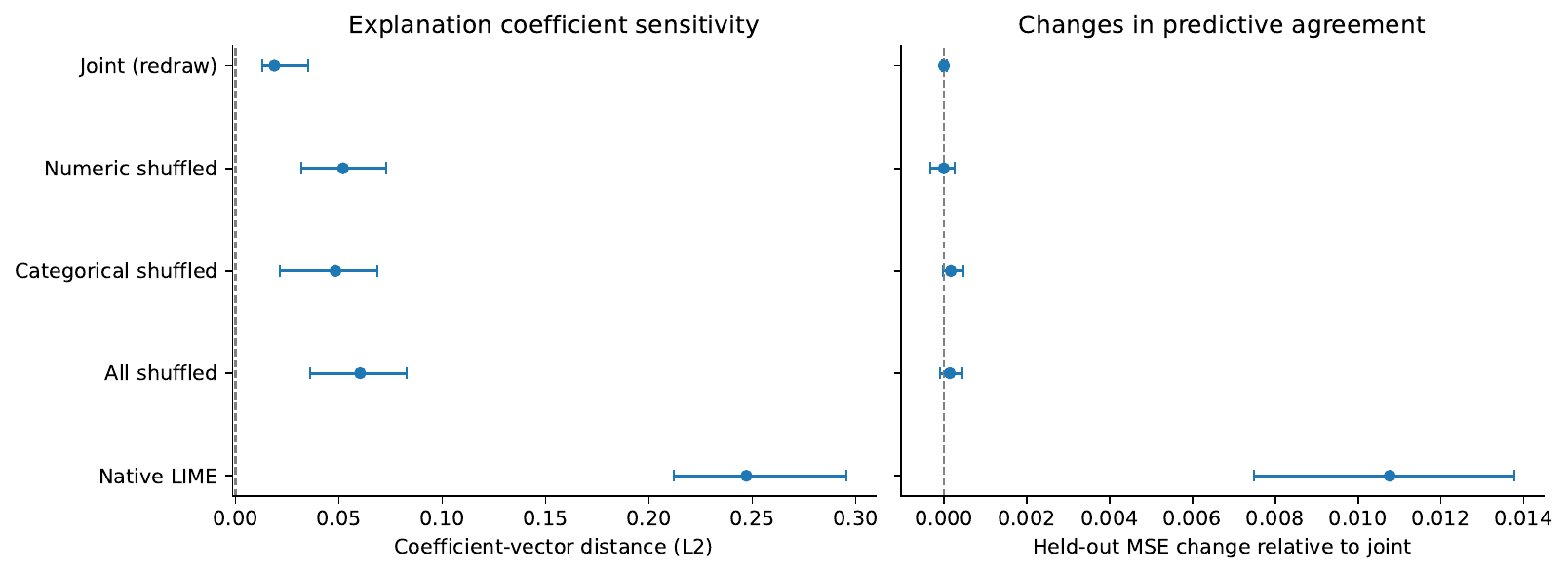}
\caption{Sensitivity to perturbation sampling. Left: mean Euclidean differences between local surrogate coefficient vectors and the joint-reference explanation. Right: mean paired changes in held-out squared surrogate-output disagreement; whiskers span the minimum and maximum split-level mean difference across ten data splits. The native LIME arm is a broader configuration comparison, not an isolated test of feature independence.}
\label{fig:attribution_sensitivity}
\end{figure*}

The third experiment compared six sampling configurations across 250 targets from ten Adult data splits (Figure~\ref{fig:attribution_sensitivity}; Table~\ref{tab:adult_lime_confirmation}). For each target, all configurations shared the target model and held-out evaluation observations.

\begin{table*}[t]
\centering
\caption{Adult/LIME results for 250 explanation targets across ten randomized data splits. Values are target-level means from the consolidated run. MSE denotes held-out squared surrogate-output disagreement; exact error is absolute disagreement at the explanation target. Coefficient differences and top-feature changes are relative to the joint reference. Flagged is the anomaly rate; Outside is the fraction of held-out surrogate outputs outside $[0,1]$.}
\label{tab:adult_lime_confirmation}
\small
\setlength{\tabcolsep}{6.0pt}
\begin{tabular}{lrrrrrr}
\toprule
Sampler
& MSE
& Exact error
& $\|\Delta\beta\|_2$
& Top change (\%)
& Flagged (\%)
& Outside (\%) \\
\midrule
Joint reference
& 0.00642 & 0.04280 & 0.00000 & 0.0 & 4.20 & 3.55 \\
Joint redraw
& 0.00642 & 0.04376 & 0.01884 & 8.0 & 4.23 & 3.59 \\
Numerical shuffle
& 0.00641 & 0.03385 & 0.05203 & 16.4 & 4.37 & 3.87 \\
Categorical shuffle
& 0.00658 & 0.03391 & 0.04839 & 17.6 & 4.78 & 4.05 \\
All-feature shuffle
& 0.00656 & 0.03350 & 0.06033 & 20.4 & 4.81 & 3.83 \\
Native LIME
& 0.01718 & 0.08949 & 0.24718 & 90.0 & 35.90 & 18.08 \\
\bottomrule
\end{tabular}
\end{table*}

Shuffling changed the surrogate coefficient vectors: their mean distances from the joint reference were 0.05203 for numerical features, 0.04839 for categorical features, and 0.06033 for all features. The top-ranked feature changed in 16.4\%, 17.6\%, and 20.4\% of those evaluations, respectively. Yet the five empirical configurations had similar mean held-out squared disagreement, ranging from 0.00641 to 0.00658. Changes in coefficients were therefore more pronounced than changes in predictive fidelity.

Native LIME differed more sharply: its mean coefficient distance was 0.24718, and the top-ranked feature changed in 90.0\% of evaluations. Mean held-out squared disagreement was 0.01718 versus 0.00642 for joint sampling, and mean exact-target absolute error was 0.08949 versus 0.04280. The anomaly detector flagged 35.90\% of native LIME perturbations, compared with 4.20\% for joint sampling, 4.81\% for all-feature shuffling, and 10.10\% for independent held-out real observations.

The mean paired increase in held-out squared disagreement for native LIME relative to joint sampling was 0.01077. This difference was positive in all ten splits; the split-level mean differences ranged from 0.00748 to 0.01378. These repeated splits reuse observations from the same Adult dataset and should not be interpreted as independent datasets.

The Ridge surrogates sometimes produced outputs outside $[0,1]$: 18.08\% of held-out outputs under native LIME sampling and 3.55\% under joint resampling. These raw regression outputs are not always valid probability forecasts, so we report squared surrogate-output disagreement rather than calling the metric a proper probabilistic score. Restricting outputs to $[0,1]$ would resolve this particular range violation but would not by itself demonstrate calibration against observed outcomes.

Because native LIME differed from the empirical generators in both sampling scale and construction, the comparison does not isolate one cause. Across the controlled procedures, however, coefficient changes were not matched by comparable changes in held-out fidelity. Attribution sensitivity and predictive agreement therefore describe distinct properties of an explanation.

\section{Discussion}

The experiments show that a local surrogate's average agreement with its target does not fully characterize the explanation. Error may vary across the neighborhood, perturbations may not represent the intended inputs, and the explanation may change when its sampling procedure changes. We therefore distinguish three diagnostics: fidelity to the target model, validity of the perturbations, and sensitivity of the fitted explanation. Each answers a different question; none alone establishes reliability for a particular purpose.

\subsection{Fidelity Depends on the Evaluation Distribution}

Experiment 1 illustrates why a neighborhood-wide score may not describe performance near an explanation target. Most sampled observations fell farther away, where the logistic surrogate agreed more closely with its target. This did not make the aggregate score wrong; it accurately summarized performance under its chosen evaluation distribution. Problems arise when readers interpret that average as evidence about a smaller, differently weighted region. The distinction between the fitting distribution \(Q_{x_0}\) and evaluation distribution \(\mu_{x_0}\) therefore matters: the same surrogate may receive different scores under different evaluation distributions.

The fidelity loss measures absolute disagreement with the target under a specified distribution; TMS additionally shows whether that disagreement improves upon an identified reference predictor. Experiment~1 yielded markedly lower TMS near the explanation target than across the full neighborhood. Neither measure establishes real-world calibration or epistemic uncertainty. We therefore recommend retaining aggregate fidelity summaries while also examining the target input and any regions relevant to the explanation's intended use.

\subsection{Perturbation Validity Is a Separate Problem}

Experiment 2 distinguishes two sampling problems. Independent sampling can disrupt observed feature relationships, producing many low-density observations when correlations are strong. Preserving covariance reduced this mismatch in our tests but did not enforce input constraints. Both Gaussian generators also assigned substantial probability to negative values of necessarily nonnegative Adult variables. These are distinct problems: one concerns the reference distribution, the other the admissible input domain.

We use \textit{probability leakage} for probability assigned outside explicit constraints, not simply to rare inputs. Neither leakage nor low density proves that a surrogate is inaccurate. A method may deliberately explore hypothetical or out-of-distribution inputs, for instance to test robustness. The question is whether its fitting and evaluation distributions match the explanatory task. An explanation of behavior among plausible records should not silently substitute an unconstrained population; a study of hypothetical inputs should state that goal.

\subsection{Attribution Sensitivity Does Not Necessarily Imply Poor Fidelity}

Experiment 3 shows that sampling choices can alter coefficients and feature rankings without producing comparable changes in held-out fidelity. Two surrogates can approximate the same target similarly across an evaluation neighborhood yet assign different coefficients to features. Neither set is automatically incorrect: when a target contains nonlinearities or interactions, locally weighted linear models can describe different aspects of its behavior depending on the inputs and weights used for fitting.

The native LIME configuration showed larger coefficient differences, greater held-out disagreement, and higher anomaly rates across the ten splits. But it differed from the empirical configurations in more than one sampling choice, so we cannot assign those differences to a single mechanism or claim that one generator is always preferable. Ridge regression also produced outputs outside \([0,1]\), particularly with native LIME sampling. Squared output disagreement remains defined, but those outputs are not valid probability forecasts. Fidelity and attribution sensitivity must therefore be interpreted as complementary measurements.

\subsection{Implications for Model-Agnostic Evaluation}

Our evaluation is model-agnostic with respect to the target: it needs only target predictions and a queryable local surrogate, not the target's internals. The same principles can therefore apply to predictive surrogates of neural networks, tree ensembles, and other models. LIME is one implementation, not a requirement. The scope is narrower than all of XAI, however; attribution-only methods without a predictive surrogate may need different fidelity measures.

A practical evaluation should first specify the intended input distribution. It should then score the surrogate on observations independent of fitting, both across the neighborhood and near the target. Constraint checks can measure inadmissible inputs; distributional checks can identify unusual combinations without calling them impossible. Repeated explanations under controlled sampling changes can reveal attribution sensitivity. We recommend reporting these diagnostics separately: a composite score would require application-specific judgments about their relative importance.

Poor local fidelity, invalid perturbations, or substantial coefficient changes each give practitioners a different reason to question an explanation. None supplies a universal threshold for accepting or rejecting it; acceptable error depends on the application and its consequences. Moreover, a surrogate can faithfully reproduce a flawed target model. Target-model fidelity measures agreement with that model, and TMS measures improvement over a stated reference; neither establishes whether the target is accurate or suitable for deployment. The aim is not to certify an explanation as trustworthy, but to make its approximation, assumptions, and limitations easier to assess.

\section{Limitations and Future Work}

\subsection{Limitations}

Our experiments use binary classification, synthetic and tabular data, and a limited set of target models, surrogates, and perturbation procedures. The mathematical evaluation does not require a particular target architecture, but we have not tested every explanation method or data type. The Adult fidelity tests also use nearby empirical observations as their evaluation distribution. This suits explanations about plausible records, but may not suit hypothetical interventions or out-of-distribution behavior.

Density estimates and anomaly flags depend on feature representation, parameter choices, and reference data; they identify unusual observations, not impossible ones. Constraint checks identify inadmissible inputs only relative to the stated rules. The attribution experiment compares linear surrogates within one implementation, and its native LIME arm differs from the empirical arms in several respects. Its Ridge surrogates sometimes produce outputs outside \([0,1]\), so we report squared output disagreement rather than treating these outputs as valid probability forecasts. Constraining the range alone would not establish calibration. The ten Adult splits also reuse the same dataset; they test robustness across partitions and targets, not independent populations.

\subsection{Future Work}

Future work should test how evaluation distributions affect fidelity for different explanatory objectives, including plausible-record descriptions and hypothetical feature changes. Comparing additional surrogate classes, multiclass classifiers, regression models, and other data types under common evaluation procedures would clarify which results generalize beyond our implementations.

It would also be useful to compare generators that preserve both feature dependencies and input constraints, measuring their effects on explanation quality rather than assuming that more realistic perturbations always improve fidelity. Finally, bounded probabilistic surrogates and explicit estimates of uncertainty about the fitted explanation could help separate systematic approximation error from variation across repeated fits.

\section{Conclusion}

Our experiments show that neighborhood-wide scores can conceal near-target errors, Gaussian perturbations can assign probability to inadmissible inputs, and sampling choices can alter feature attributions without comparable changes in predictive fidelity. These are distinct properties of local surrogates, not a single measure of reliability.

We evaluate target-model fidelity as disagreement with a queryable target under a specified distribution, and use TMS to report reference-relative improvement when a baseline is available. For probability-valued surrogates, squared disagreement admits an interpretation through expected Brier-score regret. This evaluation is model-agnostic with respect to target architecture, but perturbation validity and explanation stability require separate checks. The goal is not to certify an explanation as trustworthy, but to identify what it approximates, how closely it does so, and which assumptions affect its interpretation.
\\

\noindent \textbf{Data Availability Statement}: All materials used to generate the results in this paper are openly available in a public repository: \textbf{\small \url{https://github.com/christopherburger/TMS}}


\printbibliography

\end{document}